\documentclass[runningheads]{llncs}

\usepackage{eccv}

\usepackage{eccvabbrv}

\usepackage{graphicx}
\usepackage{booktabs}

\usepackage[accsupp]{axessibility}  

\usepackage[pagebackref,breaklinks,colorlinks,citecolor=eccvblue]{hyperref}
\usepackage{hyperref}

\usepackage{orcidlink}

\usepackage{multirow}

\begin{document}

\title{Can Vision-Language Models Assess Proxemic Risk from Egocentric Robot Images?}

\titlerunning{Vision-Language Models for Proxemic Risk Assessment}

\author{
Vladyslava Rudas\inst{1} \and
Dmytro Kuzmenko\inst{2,3} 
}

\authorrunning{V. Rudas and D. Kuzmenko}

\institute{
Department of Computer Science,\\
National University of Kyiv-Mohyla Academy, Kyiv, Ukraine\\
\email{v.rudas@ukma.edu.ua}
\and
Department of Multimedia Systems,\\
National University of Kyiv-Mohyla Academy, Kyiv, Ukraine\\
\email{kuzmenko@ukma.edu.ua}
\and
Department of Computer Science,\\
University of Turin, Turin, Italy\\
\email{dmytro.kuzmenko@unito.it}
}


\maketitle

\begin{abstract}
  Assessing proxemic danger from a robot's egocentric perspective is critical for safe embodied navigation in human environments and requires both visual and contextual reasoning. We evaluate three opensource vision-language models (VLMs) (\textit{InternVL}, \textit{Qwen-VL}, and \textit{SmolVLM}) on the classification of egocentric robot images into four danger levels, comparing three prompting strategies and two rounds of QLoRA fine-tuning against a stratified random baseline. Without fine-tuning, all models perform near the baseline, while fine-tuning yields only modest overall improvements. However, \textit{Qwen-VL} with an advanced prompt achieves substantially higher recall for high-danger cases than the other models. An analysis of person localization further shows that correct danger classification does not correspond to better spatial grounding, indicating that a model may produce a useful safety label without attending to the relevant region of the scene. These results show that current VLMs remain limited in fine-grained proxemic reasoning and spatial grounding, although targeted prompting and fine-tuning can improve high-danger detection in selected models.
\end{abstract}

\section{Introduction}
\label{sec:intro}

Mobile robots operating in environments shared with humans must constantly assess proxemic risk to navigate safely and avoid causing discomfort or collisions. A robot that is unable to correctly assess a crowded area, specifically, a high level of danger, or cannot distinguish it from an empty corridor, may invade personal space, fail to give way, or perform dangerous manoeuvres. Most existing approaches to safety assessment are based on explicit geometric reasoning or specialised depth sensors, and the question \textit{"Can universal visual understanding replace or complement such systems?"} remains open for further research.

Proxemics is a field of study that explores personal space, formalized by Edward T. Hall \cite{hall1966hidden}. Intrusion into personal comfort and danger zones is considered inappropriate and can cause stress or physical harm; therefore, for a robot to move around safely, constant assessment of the environment is required \cite{samarakoon2022review}.

Egocentric images from the robot's onboard camera provide a cost-effective source of data for proxemic analysis: they capture the robot's immediate field of view, the relative positions of people in the vicinity, and their density. Converting raw visual data into a structured risk assessment is a key objective of this study, since, as the robot is an embodied agent, this result must reflect its actual spatial understanding of the scene.

Vision-language models (VLMs) have demonstrated highly consistent performance in zero-shot and few-shot conditions when performing visual perception tasks; therefore, they may be an effective approach to solving this problem. Also, vision-language models can be manipulated by adjusting the wording of prompts and lightweight fine-tuning for specific tasks, without the need for full retraining. However, \textit{"Will this flexibility allow for reliable reasoning regarding proxemic risk?"} remains an open question requiring further research.

For this purpose, we create a new dataset on proxemic danger based on the JRDB \cite{martinmartin2023jrdb}. Our dataset is categorized into classes based on levels of proxemic danger, namely: high, moderate, low, and minimum, for 1,243 images. We will publish the modified dataset as an open-source artifact to ensure reproducibility and facilitate future comparative testing. We test three VLM models with open weights for their effectiveness in the risk assessment task, using three prompting methods and fine-tuning QLoRA in two rounds and we also evaluate whether each model's predictions are spatially grounded in the correct part of the scene.

\section{Related Work}
\subsection{Proxemics and Human-Robot Interaction Safety}
The concept of proxemics, as the study of human interpersonal space and personal boundaries, was formulated by Edward T. Hall \cite{hall1966hidden}, who identified four zones of distance: intimate, personal, social, and public. In robotics, compliance with these boundaries is a recognised requirement for safe, socially acceptable navigation \cite{samarakoon2022review, patompak2019proxemics}. Daza et al. \cite{daza2021social} proposed an approach to crowd navigation based on these principles, describing the impact of personal comfort zones on comfortable robot movement within a crowd. 

Our work applies Edward T. Hall’s concept of proxemic zones as a classification criterion, with these zones being treated as levels of danger. We aim to find out whether VLMs can identify the correct zone based solely on an egocentric image, without the use of explicit distance sensors.

\subsection{Egocentric Vision in Robotics}
Perception from an egocentric perspective is defined by reasoning from the agent's own first-person viewpoint, which ensures a natural correspondence between the input and the action the agent must take. In their study, Martin-Martin et al. \cite{martinmartin2023jrdb} presented the JRDB, a massive dataset consisting of egocentric observations made by a robot under various conditions.

We use the egocentric setting as an input channel, applying it to the task of classifying proxemic danger zones.

\subsection{Vision-Language Models for Scene Understanding}
The development of VLMs, from the problem of image captioning \cite{vinyals2015show} to the present day, reflects a fundamental shift: early systems mapped images to a fixed vocabulary, whereas transformer-based architectures \cite{vaswani2017attention, dosovitskiy2020image} enabled cross-modal reasoning between visual and linguistic representations. The key step in developing VLMs was the introduction of CLIP \cite{radford2021clip}, which demonstrated that contrastive pre-training on image-text pairs provides transferable visual representations aligned with natural language, while Flamingo \cite{alayrac2022flamingo} went further and introduced few-shot multimodal reasoning. Instruction tuning, applied to vision in LLaVA by Liu et al. \cite{liu2023visual}, made it possible to solve the open-ended problems of our benchmark.

We select three open-weight models and test whether their spatial reasoning is sufficient for proxemic classification. This evaluation is motivated by the broader limitations of VLM spatial intelligence identified by Yu et al. \cite{yu2025spatial}, but, to the best of our knowledge, has not previously been examined from the egocentric perspective of a robot operating in human environments.

\subsection{Prompt Engineering and Fine-Tuning }
Prompt formulation strongly affects VLM behaviour: Kojima et al. \cite{kojima2022zeroshot} demonstrated that chain-of-thought reasoning can be induced solely through prompts, and Ge et al. \cite{ge2023cot} applied chain-of-thought prompt tuning specifically to VLMs, while Lee et al. in their study \cite{lee2026formattax} showed that certain choices regarding prompt formatting can potentially cause measurable accuracy losses, the so-called \textit{formatting tax}, this is why we compare three structurally distinct prompt strategies.

To achieve effective parameter adaptation, we use QLoRA \cite{dettmers2023qlora}, which enables the fine-tuning of quantised large models with minimal computational resources.

In our work, we examine three prompt strategies to assess whether this helps, harms, or has no effect on the results. We also conducted two rounds of fine-tuning to observe how the models respond to training duration and the number of layers adapted.

\section{Methodology}
\subsection{Dataset}
The proposed dataset consists of 1,243 egocentric images from a mobile robot in realistic environments, split into indoor (with $n = 713$: good lighting $n = 505$, poor lighting $n = 208$) and outdoor (with $n = 530$). This distinction was made to account for the practical impact of lighting in real-world conditions.

Each image was labeled with one of four danger levels, following Hall's proxemic zones \cite{hall1966hidden}: \textbf{High} ($n = 222$, 17.9\%) – the person is within the close personal space; without immediate action, a collision and/or discomfort is unavoidable. \textbf{Moderate} ($n = 407$, 32.7\%) – the person is within the personal or close social space; trajectory adjustment is required. \textbf{Low} ($n = 299$, 24.1\%) – a person is present but at a safe social distance; required monitoring. \textbf{Minimum} ($n = 315$, 25.3\%) – no proxemic risk. The annotation was carried out manually rather than on the basis of precise distance thresholds, which could potentially lead to inaccuracies.

Class imbalance reflects the real occurrence rate: high-danger situations occur least frequently, but detecting them is the most costly; this is why we focus on the recall rate for the high-danger class alongside the overall weighted F1 score.

\subsection{Models}
We select three VLMs with open weights and a suitable size for local deployment.

\textbf{Qwen2.5-VL-3B-Instruct} \cite{bai2023qwenvl} is a model from Alibaba with 3 billion parameters, designed to work with instructions. It utilises a dynamic-resolution visual encoder and a language foundation based on the Qwen2.5 architecture, which supports detailed spatial understanding and the generation of structured output.\footnote{\url{https://huggingface.co/Qwen/Qwen2.5-VL-3B-Instruct}}

\textbf{InternVL3.5-4B} \cite{chen2024internvl} is a 4-billion-parameter model from the InternVL family, developed by OpenGVLab. It utilises a high-resolution visual encoder trained using contrastive and generative objectives, and has demonstrated high performance in spatial reasoning and visual question-answering tasks.\footnote{\url{https://huggingface.co/OpenGVLab/InternVL3_5-4B}}

\textbf{SmolVLM2-2.2B-Instruct} \cite{huggingface2024smolvlm} is a 2.2 billion parameter model by HuggingFace that is suitable for efficient inference on limited hardware. This is the smallest model in our evaluation, serving as a lower bound for what lightweight VLMs can achieve on this task.\footnote{\url{https://huggingface.co/HuggingFaceTB/SmolVLM2-2.2B-Instruct}}

\subsection{Prompt Strategies}
To assess the impact of prompt structure on performance, three prompts of increasing complexity that share one set of proxemic-zone definitions and identical JSON output format were created: \textbf{(i) Simple}: a brief task description to test the models' out-of-the-box knowledge \cite{kojima2022zeroshot}; \textbf{(ii) Moderate}: added a set of instructions and defined output constraints; and \textbf{(iii) Advanced}: an internal chain of reasoning \cite{ge2023cot} based on visual cues, which closely resembles instruction-based reasoning chains \cite{zhang2023instruction}. Each requires a JSON response with a distance zone, distance estimate, bounding box, and overall danger level; only the \textit{danger level is scored for classification}, and the \textit{bounding box is scored separately for grounding}.

\subsection{Fine-Tuning Procedure}
We perform fine-tuning with parameter-efficient QLoRA in two stages, creating three checkpoints: \textit{before}, \textit{after}, and \textit{after\_2}. A total of 200 images were selected for the training set, distributed evenly across danger levels. To avoid data leakage, images were taken from videos different from those used for creating the test set. \textbf{Stage 1}: adapts attention projections only (q\_{proj}, v\_{proj}). \textbf{Stage 2}: extends this to all main projections (q\_proj, k\_proj, v\_proj, o\_proj, gate\_proj, up\_proj, down\_proj, lm\_head). Both stages use \textit{r = 8, $\alpha$ = 16, dropout = 0.05}, and no offset adaptation. \textit{Stage 2} extends the coverage of the adapter to improve task consistency, whilst the two-stage configuration allows us to assess whether additional fine-tuning improves performance or causes regression.

\subsection{Evaluation Metrics}
We report the following metrics for each model $\times$ fine-tuning stage $\times$ prompt type configuration:

\textbf{Accuracy} – the fraction of correctly classified images.

\textbf{Weighted F1} – the average F1 score across all classes, weighted by the frequency of occurrence, which reflects the overall quality of the classification under class imbalance.

\textbf{Recall of high danger class} – recall specifically for the \textit{high} danger class, defined as the proportion of correctly identified scenes with high danger for the robot. This is a key safety metric, as a false negative for a scene with high danger carries a greater cost than any other classification error.

\textbf{IoU} – a correctly classified danger level may not mean that the model is seeing the right person. We also measure how well the predicted bounding box matches the ground truth person position using IoU, with mean IoU, ratio of IoU $>$ 0.5, and ratio of IoU = 0. This helps us to understand if the model knows \textit{where} the scene is dangerous.

\section{Results}
\subsection{Overall Performance}
Table ~\ref{tab:overall} summarizes accuracy, weighted F1, and high-danger recall by model and fine-tuning stage (random baseline: wF1 = 0.25).

\begin{table}[h]
\centering
\caption{Overall performance by model and fine-tuning stage
(averaged over prompt strategies).}
\label{tab:overall}
\begin{tabular}{llccc}
\toprule
\textbf{Model} & \textbf{Stage} & \textbf{Acc.} & \textbf{wF1} &
\textbf{High Recall} \\
\midrule
Random baseline & — & 0.25 & 0.25 & 0.25 \\
\midrule
InternVL  & before   & \textbf{0.304}& 0.228 & 0.026 \\
InternVL  & after    & 0.300 & 0.230 & 0.033 \\
Qwen      & before   & 0.280 & 0.219 & 0.395 \\
Qwen      & after    & 0.254 & 0.211 & 0.432 \\
Qwen      & after\_2 & 0.276 & 0.234 & \textbf{0.440}\\
SmolVLM   & before   & 0.288 & 0.212 & 0.016 \\
SmolVLM   & after    & 0.289 & \textbf{0.238} & 0.069 \\
SmolVLM   & after\_2 & 0.274 & 0.227 & 0.055\\
\bottomrule
\end{tabular}
\end{table}

As can be seen, all of the models examined show results that are close to the random baseline (weighted F1 score \textit{(wF1)} 0.21–0.24). However, by shifting the focus to the study of high levels of danger, better results can be observed; specifically, for the Qwen model, the recall reaches  0.4 at all stages, versus near-zero for InternVL and SmolVLM. Fine-tuning does not lead to any gain ($\Delta$wF1 $<$ 0.02).

Table~\ref{tab:prompt} gives classification and grounding by prompt strategy. 
The simple prompt achieves the highest overall F1 score and mean IoU, whilst the advanced prompt with a chain of reasoning yields the highest recall for the class of high danger. This inversion may point to a "format tax" \cite{lee2026formattax}: more complex prompts introduce structured reasoning, which improves the detection of outliers but at the expense of overall classification consistency. The moderate prompt shows no clear advantage in either of these objectives; however, the use of this prompt yields the lowest total misalignment score.
\begin{table}[h]
\centering
\caption{Performance by prompt strategy.}
\label{tab:prompt}
\begin{tabular}{lcccc}
\toprule
\textbf{Prompt} & \textbf{wF1} & \textbf{High Recall} & \textbf{Mean IoU} & \textbf{IoU = 0} \\
\midrule
Simple   & \textbf{0.303} & 0.122 & \textbf{0.232} & 0.434 \\
Moderate & 0.263 & 0.196 & 0.110 & \textbf{0.416} \\
Advanced & 0.249 & \textbf{0.316} & 0.196 & 0.451 \\
\bottomrule
\end{tabular}
\end{table}

\subsection{Spatial Grounding}
Tab \ref{tab:model_performance} reports grounding by model and fine-tuning stage. It is evident from the table that the only model providing useful localisation is Qwen: it demonstrates relatively high metric values compared to other models, namely an average IoU of up to 0.46 and IoU $>$ 0.5 on over half of the predictions before fine-tuning. The other two models generally do not provide localisation. As with classification, fine-tuning does not improve localisation for any of the models; in fact, it even worsens it slightly.
\begin{table}[h]
\centering
\caption{Grounding performance by model and fine-tuning stage.}
\label{tab:model_performance}
\begin{tabular}{llccc}
\toprule
\textbf{Model} & \textbf{Stage} & \textbf{Mean IoU} & \textbf{IoU $>$ 0.5} & \textbf{IoU = 0} \\
\midrule
InternVL & before   & 0.127 & 0.023 & 0.404 \\
InternVL & after    & 0.124 & 0.031 & 0.426 \\
\midrule
Qwen     & before   & \textbf{0.459} & \textbf{0.545} & \textbf{0.216} \\
Qwen     & after    & 0.413 & 0.481 & 0.236 \\
Qwen     & after\_2  & 0.419 & 0.482 & 0.230 \\
\midrule
SmolVLM  & before   & 0.061 & 0.001 & 0.612 \\
SmolVLM  & after    & 0.046 & 0.004 & 0.680 \\
SmolVLM  & after\_2  & 0.051 & 0.005 & 0.667 \\
\bottomrule
\end{tabular}
\end{table}

\subsection{Per-Class Analysis}
Tab ~\ref{tab:perclass} reports precision, recall, and F1 per danger class for the best overall configuration (Qwen, after\_2, simple prompt) and the best high-recall configuration (Qwen, after, advanced prompt), alongside mean IoU per true class, showing that grounding quality does not track classification performance on a per-class basis either.
\begin{table}[h]
\centering
\caption{Per-class metrics for the two best Qwen configurations. Precision (P), recall (R), F1-score (F1), and mean IoU (grounding accuracy on that class) are reported for each class.}
\label{tab:perclass}
\begin{tabular}{llcccc}
\toprule
\textbf{Config} & \textbf{Class} & \textbf{P} & \textbf{R} & \textbf{F1} & \textbf{IoU} \\
\midrule
\multirow{4}{*}{Qwen / after\_2 / sim}
 & high     & 0.276 & 0.279 & 0.278 & \textbf{0.505}\\
 & moderate & \textbf{0.378}& 0.231 & 0.287 & 0.502 \\
 & low      & 0.290 & 0.098 & 0.146 & 0.473 \\
 & minimum  & 0.320 & \textbf{0.700}& \textbf{0.440}& 0.382 \\
\midrule
\multirow{4}{*}{Qwen / after / adv}
 & high     & 0.182 & \textbf{0.790}& \textbf{0.296}& \textbf{0.442}\\
 & moderate & 0.301 & 0.100 & 0.150 & 0.409 \\
 & low      & \textbf{0.316}& 0.020 & 0.038 & 0.371 \\
 & minimum  & 0.210 & 0.100 & 0.135 & 0.327 \\
\bottomrule
\end{tabular}
\end{table}
\section{Discussion}
\subsection{Which Configuration Is Actually Deployable?}
No single configuration allows for the optimization of both overall accuracy and high-danger recall at the same time; however, in order to integrate the model into the robot control system, it must serve as a reliable danger trigger. The two best-performing combinations investigated: \textit{(1)} Qwen + simple prompt + a second extended round of fine-tuning, provides the best overall performance (weighted F1 = 0.290) and balanced classification across all danger levels; \textit{(2)} Qwen + advanced prompt + first simple fine-tuning round achieves the highest recalls for high danger (0.79), but performs poorly on other classes. It is also important to note that this high-danger configuration is characterised by a high recall rate but very low precision and insufficient effectiveness for the other classes. Therefore, its practical deployability as a safety system has not yet been conclusively demonstrated. In contrast, InternVL and SmolVLM cannot reliably detect situations of high danger even after fine-tuning.

\subsection{Classification Success Does Not Imply Spatial Grounding}
Our research shows that the models examined generally do not take into account data on the spatial location of people when determining the level of danger. This conclusion is supported by the IoU value, which shows no notable difference between correctly and incorrectly classified frames (0.204 versus 0.215), and even Qwen’s best performance on true positives in the high danger class (IoU = 0.488) barely exceeds its performance on false negatives in the same class (IoU = 0.464). This suggests that a high danger classification label does not confirm that the model has identified the responsible person for this situation.

\subsection{Why Fine-Tuning Yields Limited Gains}
The results suggest that the QLoRA configuration offers minor improvement, which is may be due to structural limitations rather than the choice of tuning parameters.

First, the dataset is likely too small; fine-tuning was performed on a balanced set of 200 images, meaning each class had exactly 50 examples, which could potentially limit the learning signal.

Second, Qwen already demonstrates a high recall rate for high danger class without training, suggesting that spatial reasoning is largely acquired during pre-training rather than in the layers adapted for QLoRA.

Third, additional fine-tuning is detrimental to small models, such as SmolVLM (2.2 billion parameters). This is due to the phenomenon known as "catastrophic forgetting" \cite{kirkpatrick2017catastrophicforgetting}, which arises from the difficulty of adapting new, task-specific patterns to the model’s architecture when capacity is limited.

\section{Conclusion}
Our work explored the ability of vision-language models to predict the danger level in egocentric robot pictures in terms of proxemics and to provide enough spatial grounding for the purposes of an embodied control loop. The answer to the question we posed in the title of the paper is \textit{only partially and unevenly can VLMs assess proxemic risk from egocentric data}.

One of the models, Qwen, demonstrates the ability to detect high levels of danger; however, the other two models produce almost entirely zero results when working with the same class, regardless of prompts or fine-tuning. All configurations return overall F1-scores close to the random baseline. Grounding accuracy follows the same pattern: Qwen is the only model whose predictions regarding the bounding box closely match the ground truth -- but, most importantly, this alignment is largely independent of classification accuracy: even Qwen’s correct predictions of a high level of danger do not demonstrate reliably better agreement than those which the model failed to detect. From these results, we can conclude that four-class proxemic reasoning based on single-view egocentric frames remains an unsolved problem for modern VLMs.

Future research should explore the use of larger training datasets and comparisons with more powerful API-scale models to determine whether performance limitations are due to the model’s architecture or depend on its scale, and whether classification and grounding can be jointly improved.
\bibliographystyle{splncs04}
\bibliography{main}
\end{document}